\documentclass{article}

\usepackage[preprint]{neurips_2026}

\usepackage[utf8]{inputenc}
\usepackage[T1]{fontenc}
\usepackage{hyperref}
\usepackage{url}
\usepackage{booktabs}
\usepackage{amsfonts}
\usepackage{nicefrac}
\usepackage{microtype}
\usepackage{xcolor}
\usepackage{graphicx}
\usepackage{subcaption}
\usepackage{xurl}

\usepackage{amsmath}
\usepackage{amssymb}
\usepackage{mathtools}
\usepackage{amsthm}

\usepackage[capitalize,noabbrev]{cleveref}

\usepackage{algorithm}
\usepackage{algorithmic}

\theoremstyle{plain}
\newtheorem{theorem}{Theorem}[section]
\newtheorem{proposition}[theorem]{Proposition}
\newtheorem{lemma}[theorem]{Lemma}
\newtheorem{corollary}[theorem]{Corollary}
\theoremstyle{definition}
\newtheorem{definition}[theorem]{Definition}

\theoremstyle{remark}
\newtheorem{remark}[theorem]{Remark}

\usepackage[textsize=tiny]{todonotes}

\title{Analytical Provisioning for Attention–FFN Disaggregated LLM Serving under Stochastic Workloads}

\author{%
  Chendong Song\thanks{Corresponding author: \texttt{songcd@ust.hk}} \\
  Dept.\ of Industrial Engineering\\and Decision Analytics\\
  HKUST, Hong Kong, China\\
  \And
  Meixuan Wang \\
  Huawei Hong Kong\\Research Center, China\\
  \And
  Hang Zhou \\
   Dept.\ of Industrial Engineering\\and Decision Analytics\\
  HKUST, Hong Kong, China\\
  \And
  Hong Liang \\
  Huawei Hong Kong\\Research Center, China\\
  \And
  Yuan Lyu \\
  Huawei Hong Kong\\Research Center, China\\
  \And
  Zixi Chen \\
  Huawei Hong Kong\\Research Center, China\\
  \And
  Yuwei Fan \\
  Huawei Hong Kong\\Research Center, China\\
  \And
  Zijie Zhou\thanks{Corresponding author: \texttt{jerryzhou@ust.hk}} \\
  Dept.\ of Industrial Engineering\\and Decision Analytics\\
  HKUST, Hong Kong, China\\
}

\begin{document}

\maketitle

\begin{abstract}
Attention–FFN disaggregation (AFD) is an emerging architecture for LLM decoding that separates state-heavy, KV-cache–dominated Attention computation from stateless, compute-intensive FFN computation, connected by per-step communication. While AFD enables independent scaling of memory and compute resources, its performance is highly sensitive to the Attention/FFN provisioning ratio: mis-sizing induces step-level blocking and costly device idle time. We develop an analytical provisioning framework for AFD bundles in an $r$A--$1$F topology under stochastic workloads. Two sources of randomness shape the problem: per-slot Attention workload evolves as KV caches grow and completed requests are replenished with random prompt and decode lengths, and synchronized execution across Attention workers introduces a barrier governed by the slowest worker. We address both via a renewal-reward characterization of the per-slot stationary token load, identifying a single workload statistic $\theta$ that governs provisioning under arbitrary prefill-decode distributions and admits a nonparametric estimator from request traces. The analysis yields a closed-form mean-field rule for the optimal A/F ratio decomposing into Attention-, communication-, and FFN-bottleneck regimes, together with a Gaussian barrier-aware refinement that quantifies cross-worker synchronization overhead. A trace-calibrated AFD simulator supports the framework across workloads: the predicted optimal ratio matches the simulation-optimal within 10\%. Together, these results provide a compact, calibratable account of how stochastic workload structure determines provisioning in disaggregated LLM serving.
\end{abstract}

\section{Introduction}\label{sec:intro}
The explosive growth of large language models (LLMs) \citep{brown2020language,chowdhery2023palm,openai2023gpt,kaplan2020scaling}  has created unprecedented challenges for efficient inference serving at scale. As model sizes expand into hundreds of billions of parameters, modern LLM serving inevitably requires distributed multi-device architectures. This reality has spurred a shift from monolithic serving paradigms toward increasingly sophisticated disaggregation strategies that decompose the inference pipeline across heterogeneous hardware resources.

Early efforts of disaggregate serving focused on stage-level separation, exemplified by the Prefill-Decode (PD) disaggregation \citep{zhong2024distserve, patel2024splitwise}. LLM inference comprises two phases with distinct resource profiles: \textit{prefill} processes the input prompt and generates the KV cache via dense matrix operations (compute-bound), while \textit{decode} generates tokens autoregressively, reading the accumulated KV cache at each step (memory-bound). PD disaggregation separates these phases onto hardware matched to their respective bottlenecks. By disaggregating prefill and decode onto separate hardware pools that can be independently scaled, PD architectures enable higher decode batch sizes through request aggregation, improving GPU utilization during token generation.

Building upon this foundation, recent research has recognized that even within the decode phase, computational heterogeneity persists between different transformer components. The Attention-FFN Decoupling (AFD) \citep{wang2025step, zhang2025janus, zhu2025megascale, zuo2025serving} extends this disaggregation further: Attention layers are stateful and memory-bound (dominated by KV cache reads), whereas FFN layers are stateless and can achieve compute-bound operation with sufficient batching. Traditional coupled architectures often leave FFN compute units underutilized due to the small decode batch with TPOT constraints. AFD architecture strategically disaggregates these components, allowing multiple Attention instances to feed into an aggregated FFN instances. This topology maximizes FFN arithmetic intensity while independently scaling memory resources for Attention.

Recent work~\citep{wang2025step, zhu2025megascale, zuo2025serving} has shown that AFD can significantly improve serving efficiency by allowing Attention and FFN to scale independently. Given this flexibility, a central design decision in such systems is the ratio of Attention instances to FFN instances, which we denote as $r$. Currently, existing systems set $r$ through empirical search or naive deterministic approximations that ignore workload stochasticity. A principled analytical provisioning framework is still lacking. This gap matters because the choice of $r$ directly impacts system efficiency: when $r$ is too small, the FFN server starves for input; when $r$ is too large, Attention instances block waiting for FFN availability.

Finding a principled analytical provisioning rule, however, is non-trivial because the Attention workload is fundamentally stochastic. FFN computation depends only on batch size and remains stable across decode steps. In contrast, Attention workload evolves continuously—the KV cache grows with each step, and completed requests are replaced by new ones with variable prompt lengths.

As AFD remains an emerging paradigm with no mature open-source implementations, principled \textbf{analytical foundations} are needed to guide this design space before extensive system building begins. In this paper, we provide such foundations through a rigorous probabilistic framework that captures the stochastic dynamics of Attention-side workload and derives a closed-form expression for the optimal A/F ratio. Our simulation-based validation demonstrates that the theory accurately predicts system behavior; real-system validation is left to future work as AFD implementations mature. Our contributions are summarized as follows:

\textbf{(i) Probabilistic workload model.} We develop an analytical framework capturing the stochastic dynamics of AFD serving under continuous batching with request replenishment. Using a discrete-time renewal-reward analysis of one decode slot, we identify a single workload statistic that governs provisioning under arbitrary prompt and decode-length distributions and admits a nonparametric estimator from request traces.

\textbf{(ii) Provisioning rule with load-balancing correction.} We derive a closed-form mean-field rule for the A/F ratio that decomposes into three operating regimes (Attention-bottleneck, communication-bottleneck, FFN-bottleneck), each with an interpretable balance condition. Cross-worker synchronization stragglers are handled via a Gaussian order-statistic correction.

\textbf{(iii) Validation across workloads and hardware regimes.} We develop a trace-calibrated AFD simulator and validate the framework across diverse workload and hardware configurations. The closed-form rule matches the simulator-optimal ratio within 10\%, which is an empirical confirmation of the theory.

\section{Background on Attention-FFN Disaggregation: Pipeline and Microbatch} \label{sec:bg}

In standard LLM decoding, a monolithic architecture deploys both Attention and FFN blocks on the same hardware. To reduce kernel launch overhead and execution bubbles, practitioners typically compile the model into a static execution graph (e.g., CUDA Graphs \citep{ekelund2025boosting}), which enforces fixed batch sizes across all components. This rigidity creates a utilization problem: time per output token (TPOT) latency constraints force small batch sizes, and at low concurrency, FFN blocks cannot amortize their weight-loading cost—despite being computationally dense, they become memory-bandwidth bound rather than compute-bound.

%AFD fundamentally restructures the execution paradigm by decoupling the memory-bound Attention blocks from the compute-bound FFN blocks. By physically separating these components onto distinct hardware clusters, the architecture facilitates independent scaling and heterogeneous resource allocation. Crucially, this separation enables an asymmetric deployment ratio, where multiple stateful Attention instances feed into a single stateless FFN instance (e.g., an $r:1$ ratio). This consolidation aggregates the workload from multiple Attention streams, significantly increasing the effective concurrency for the FFN. Consequently, the FFN operation shifts from a memory-bound regime closer to the hardware's cube bound (maximum theoretical compute throughput).

AFD addresses this by disaggregating Attention and FFN onto separate hardware, enabling an asymmetric $r:1$ deployment where multiple Attention instances feed a shared FFN server. We now describe the execution mechanics of this architecture.

\begin{figure}[t]
  \centering
\includegraphics[width=0.7\linewidth]{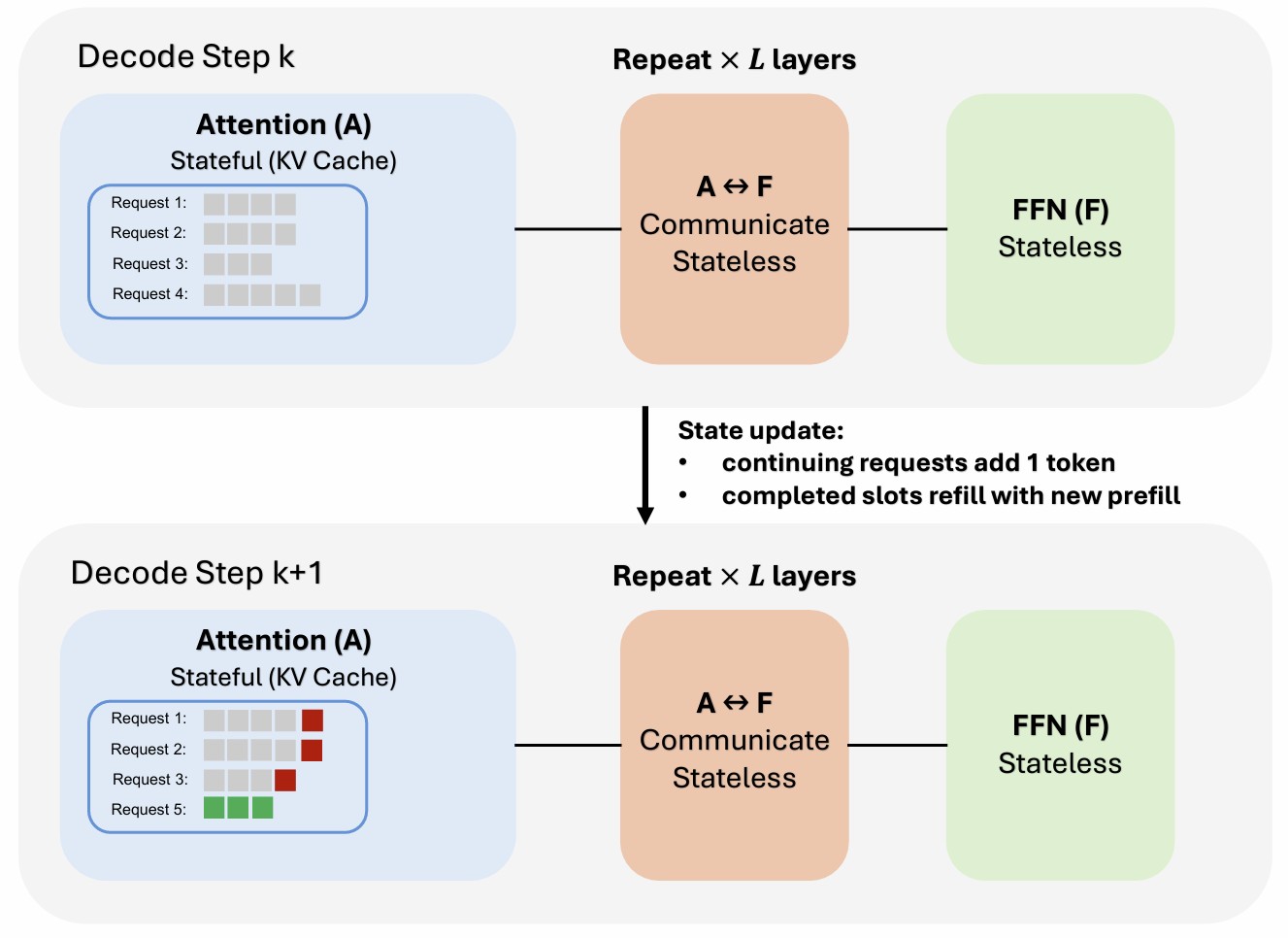}
    \caption{Architecture of AFD. Stateful Attention (A) layers manage the KV cache, while Feed-Forward Network (F) layers are stateless. During each decode step, every continuing request generates one output token whose key-value is appended to the KV cache (red blocks); when a request completes, its slot is immediately refilled with a new prefill request (green block).}
\label{fig:af}
\end{figure}

\begin{figure}[t]
  \centering
  \begin{subfigure}{0.95\linewidth}
    \centering
    \includegraphics[width=0.8\linewidth]{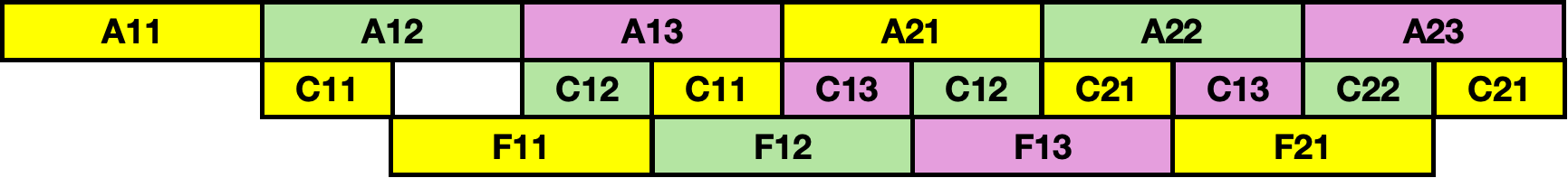}
    \caption{Ideal microbatch}
  \end{subfigure}\par\vspace{0.5em}

  \begin{subfigure}{0.95\linewidth}
    \centering
    \includegraphics[width=0.8\linewidth]{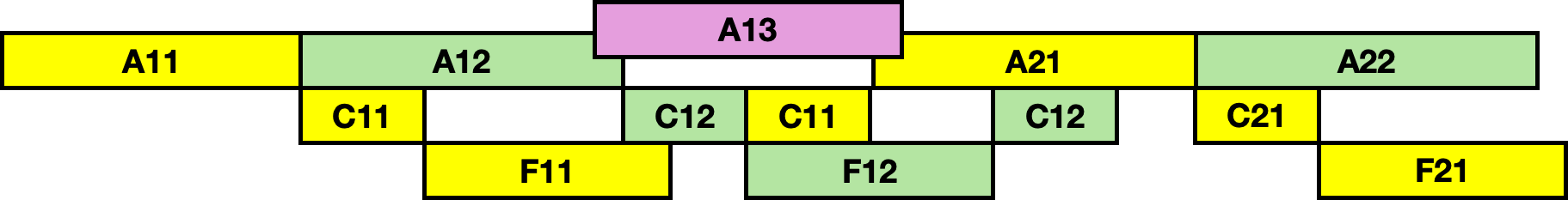}
    \caption{Issue raised after one decode step}
  \end{subfigure}
  \caption{Microbatch Pipelining and Masking. \textbf{(a)} An ideal schedule where Attention, Communication, and FFN are perfectly overlapped across microbatches, fully hiding data transfer latency. \textbf{(b)} After one decode step, Attention execution time increases (due to longer KV cache reads) while Communication and FFN remain unchanged, breaking the balanced overlap and introducing pipeline bubbles.}
  \label{fig:micro}
\end{figure}

In an AF-disaggregated deployment, each Transformer layer comprises three distinct stages: Attention computation, FFN computation, and inter-device Communication ($A \leftrightarrow F$). As illustrated in Figure~\ref{fig:af}, Attention blocks are stateful: during each decode step, every continuing request generates one output token, and its corresponding key-value is appended to the KV cache. When computing attention, the model must read the entire accumulated KV cache for each request, so the Attention cost grows linearly with sequence length. When a request completes, its slot is immediately refilled with a newly prefill request to maintain batch size (continuous batching). In contrast, FFN blocks are stateless—their computation depends only on the current activations, not on sequence history. To handle the latency introduced by transferring activations (AF communication) across $L$ layers within one decode step, microbatch pipelining is adopted, as shown in Figure~\ref{fig:micro}.

Figure~\ref{fig:micro}(a) depicts an ideal execution schedule. Here, $A_{mn}$, $C_{mn}$, and $F_{mn}$ denote Attention, Communication, and FFN operations for the $n$-th microbatch at the $m$-th layer, respectively. With sufficient microbatches (typically $\geq 3$), the system overlaps Communication and FFN of one microbatch with Attention of subsequent microbatches, thereby masking data transfer (communication) latency.

However, Attention's stateful nature introduces dynamic variability that disrupts this ideal schedule. Consider Figure~\ref{fig:micro}(b): after one decode step, assuming no request completes, every request's KV cache has grown by one token. Consequently, $A_{11}$, $A_{12}$, and $A_{13}$ all take longer than in the previous step, while $C_{mn}$ and $F_{mn}$ remain unchanged. The increased Attention time means that $A_{12}$ and $A_{13}$ can no longer fit in the gaps between layers—$A_{13}$ is deferred, leaving idle ``bubbles" in the pipeline. Depending on the workload imbalance, bubbles may appear on the Attention side (waiting for FFN) or the FFN side (starved of data from Attention). Since Attention workload drifts continuously, no static microbatch schedule remains ideal over time. This motivates our central question: how to determine the optimal Attention-to-FFN ratio ($r$:1) that minimizes expected pipeline bubbles across a long serving horizon.

\section{Mathematical Model}\label{sec:model}

We consider an AFD bundle in a general $x$A--$y$F topology, where $x$ Attention instances feed into $y$ shared FFN instances. To simplify notation, we define the Attention-to-FFN ratio $r:=x/y$ and express the topology equivalently as $r$A--$1$F. $r$ does not need to be an integer, for example, $r=3.5$ corresponds to a $7$A--$2$F configuration. Each Attention instance maintains a microbatch of $B$ requests and their associated KV caches. Decoding proceeds in synchronized steps, where each step comprises four phases:

\begin{enumerate}
    \item[(i)] \emph{Attention phase:} The $r$ Attention workers execute in parallel, each processing its microbatch of $B$ requests (total $rB$ requests across the bundle).
    \item[(ii)] \emph{A$\to$F communication:} All $r$ Attention workers transfer their intermediate activations to the shared FFN worker.
    \item[(iii)] \emph{FFN phase:} The FFN worker processes the aggregated batch of $rB$ activations.
    \item[(iv)] \emph{F$\to$A communication:} The FFN worker returns outputs to all $r$ Attention workers.
\end{enumerate}

\noindent Our goal is to determine the optimal ratio $r^*$ that maximizes system throughput.

%----------------------------------------------------------
\subsection{Latency Models}
\label{sec:latency_models}
%----------------------------------------------------------

We adopt linear latency models grounded in first-principles hardware analysis (see Appendix~\ref{app:latency} for detailed derivations) and consistent with established results in the LLM inference literature.

\noindent \textbf{Attention.}
The Attention computation reads the KV cache for all requests in the microbatch. For a microbatch of $B$ requests, let $s_b$ denote the prefill length and $i_b$ the current decode index for request $b$. The total token load is $T = \sum_{b=1}^{B} (s_b + i_b)$.
During decoding, Attention is memory-bandwidth bound—dominated by reading the KV cache from HBM \citep{kwon2023efficient,pope2023efficiently,agrawal2024taming}. By the roofline model, latency for a memory-bound operation equals data volume divided by effective bandwidth \citep{yuan2024llm,lienhart2024dissecting}, so Attention latency scales linearly with $T$: $t_A(T) = \alpha_A T + \beta_A$.

\noindent \textbf{FFN.}
The FFN processes batched activations and becomes compute-bound with sufficient batching \citep{agrawal2024taming,wei2024building}. FFN layers perform matrix multiplications whose FLOPs scale as $O(\text{batch} \times \text{hidden} \times \text{intermediate})$, yielding linear latency in batch size \citep{pope2023efficiently,chen2023dissecting}: $t_F(rB) = \alpha_F (rB) + \beta_F$.

\noindent \textbf{Communication.}
The round-trip communication cost per Attention instance is $t_C(rB) = \alpha_C (rB) + \beta_C$, where $\beta_C$ is a fixed startup cost and $\alpha_C (rB)$ is the bandwidth-dependent term. This is the standard model for distributed communication, validated extensively in GPU clusters \citep{narayanan2021efficient,li2024flash}.

%----------------------------------------------------------
\subsection{Workload Characterization}
\label{sec:workload}
%----------------------------------------------------------

\noindent \textbf{Request lengths.}
Each request has a prefill length $P \in \mathbb{Z}_{\ge 0}$ and a decode lifetime $D \in \{1, 2, \ldots\}$, where $D$ is the number of decode steps the request occupies its slot. During those $D$ steps, the decode index takes values $0, 1, \ldots, D-1$. We treat $(P_n, D_n)_{n \ge 1}$ as i.i.d.\ across requests, allowing arbitrary dependence between $P_n$ and $D_n$ within a request. We assume only that $\mathbb{E}[P]$, $\mathbb{E}[D]$, and the second moment of the per-slot stationary load (defined in Section~\ref{sec:dynamics}) are finite. No specific distributional family is required.

\noindent \textbf{Continuous batching.}
Following standard practice in LLM serving, we use continuous batching: whenever a request completes, its slot is immediately refilled with a new request from the queue, maintaining full microbatches throughout serving.

\noindent \textbf{Cross-worker load balancing.}
The $r$ Attention workers in a bundle execute synchronously, so the Attention phase waits for the slowest worker. Stochastic request dynamics induce heterogeneous token loads across workers, creating a straggler effect that grows with $r$. Implementing load-balancing routing policies~\citep{chen2026universal} can mitigate this gap, though some irreducible variance from stochastic dynamics remains.

%----------------------------------------------------------
%----------------------------------------------------------
%----------------------------------------------------------
\subsection{Optimization Objective}
\label{sec:objective}
%----------------------------------------------------------

\noindent \textbf{Cycle time.}
Let $T_j$ denote the total token load on Attention worker $j \in \{1, \ldots, r\}$. Because the $r$ workers execute synchronously, the Attention phase completes only when the slowest worker finishes. The relevant Attention-side load is therefore the \emph{barrier load} $W_{B,r} := \max_{1 \le j \le r} T_j$. The per-step cycle time is $\tau(B; r) = \max\big\{\alpha_A W_{B,r} + \beta_A,\; t_C(rB),\; t_F(rB)\big\}$.

\noindent \textbf{Throughput per instance.}
Long-run throughput is governed by the expected cycle time $\bar\tau(B; r) := \mathbb{E}[\tau(B; r)]$.
We maximize the average number of output tokens generated per unit time, normalized by the total number of instances ($r$ Attention $+$ $1$ FFN):
\begin{equation}
\mathrm{Throughput}_{\mathrm{per\text{-}inst}}(B; r) = \frac{1}{r+1} \cdot \frac{rB}{\bar\tau(B; r)},
\label{eq:throughput}
\end{equation}
where $rB$ is the number of tokens generated per cycle across the bundle.

The central challenge is that $\bar\tau(B; r)$ is shaped by two sources of stochasticity. Within each Attention worker, per-slot loads evolve as decoding progresses and slots are replenished with random prompt and decode lengths, so the workload is non-stationary at the step level even though it admits a stationary distribution under continuous batching. Across workers, the maximum in $W_{B,r}$ introduces a synchronization barrier whose expected overhead grows with $r$. The next section develops a probabilistic analysis to provide a provisioning rule.

% ============================================================
% Section 4: Analysis and Optimal A/F Ratio
% ============================================================

\section{Analysis and Optimal A/F Ratio}\label{sec:optimal}

We now derive the optimal Attention-to-FFN ratio $r^*$. The analysis must contend with two sources of randomness simultaneously: per-slot Attention workload evolves pathwise as decoding progresses, with KV caches growing and completed slots replenished by new requests of random length, and the synchronized Attention phase is governed by a barrier across the $r$ workers. Classical fork-join queueing models assume stationary, identical service times and therefore do not directly apply. We address both effects by combining a renewal-reward characterization of the per-slot stationary load with a Gaussian order-statistic correction for the cross-worker barrier. The analysis yields a closed-form mean-field provisioning rule, refined by a barrier-aware correction that captures synchronization overhead.

%----------------------------------------------------------
\subsection{Stationary Per-Slot Token Load}
\label{sec:dynamics}
%----------------------------------------------------------

By symmetry, it suffices to analyze one decode slot under continuous batching. Requests arrive at the slot sequentially. Request $n$ has prompt length $P_n \in \mathbb{Z}_{\ge 0}$ and decode lifetime $D_n \in \{1, 2, \ldots\}$, with $(P_n, D_n)_{n \ge 1}$ i.i.d.\ across requests. During its lifetime, the request occupies the slot for $D_n$ decode steps, contributing token load $P_n + a$ at age $a \in \{0, 1, \ldots, D_n - 1\}$. Let $Y$ denote the stationary token load of one slot, observed at a uniformly random decode step, and define $\theta := \mathbb{E}[Y]$, $\nu^2 := \mathrm{Var}(Y)$.

\begin{lemma}[Stationary token-load moments]\label{lem:stationary}
Assume $0 < \mathbb{E}[D] < \infty$ and
\begin{equation}
\mathbb{E}\!\left[D P^2 + P D(D-1) + \tfrac{D(D-1)(2D-1)}{6}\right] < \infty.
\label{eq:moment_assumption}
\end{equation}
Then
\begin{equation}
\theta = \frac{\mathbb{E}\!\left[D P + \tfrac{D(D-1)}{2}\right]}{\mathbb{E}[D]},
\qquad
\mathbb{E}[Y^2] = \frac{\mathbb{E}\!\left[D P^2 + P D(D-1) + \tfrac{D(D-1)(2D-1)}{6}\right]}{\mathbb{E}[D]},
\label{eq:theta_general}
\end{equation}
and $\nu^2 = \mathbb{E}[Y^2] - \theta^2$. If $P$ and $D$ are independent,
\begin{equation}
\theta = \mu_P + \frac{\mu_D - 1}{2} + \frac{\sigma_D^2}{2 \mu_D},
\label{eq:theta_indep}
\end{equation}
where $\mu_D := \mathbb{E}[D]$ and $\sigma_D^2 := \mathrm{Var}(D)$. Without independence, an additional term $\mathrm{Cov}(P, D)/\mu_D$ enters.
\end{lemma}

The proof, given in Appendix~\ref{app:lemma_stationary}, applies the discrete-time renewal-reward theorem to one decode slot, treating each request as a renewal cycle of length $D_n$.

Equation~\eqref{eq:theta_indep} identifies the workload statistic that drives provisioning: it is \emph{not} the arrival-average $\mu_P + \mu_D$ (a natural but incorrect first guess), but the stationary age-adjusted load. The decode-length variance $\sigma_D^2$ enters through length-biasing (longer requests are sampled more often when observed at a random decode step) and the prefill--decode covariance enters when long prompts induce long responses. The framework imposes no parametric assumption on $(P, D)$; Appendix~\ref{app:tracebased} gives a nonparametric estimator that calibrates $\theta$ and $\nu^2$ directly from request traces. 

\begin{remark}[Bounded request lengths]\label{rem:heavy_tail}
In practice, serving systems impose maximum context and generation lengths, so $P$ and $D$ are bounded and condition~\eqref{eq:moment_assumption} is automatically satisfied. Appendix~\ref{app:heavy_tail} discusses the role of length-biasing under unbounded distributions for completeness.
\end{remark}

%----------------------------------------------------------
\subsection{Barrier-Aware Attention Load}
\label{sec:barrier}
%----------------------------------------------------------

Each Attention worker $j \in \{1, \ldots, r\}$ holds $B$ slots with stationary loads $Y_{j,b}$, i.i.d.\ across $(j, b)$. The total load on worker $j$ is $T_j^{(B)} = \sum_{b=1}^{B} Y_{j,b}$, and the synchronization barrier load is $W_{B,r} = \max_{1 \le j \le r} T_j^{(B)}$. Throughout this subsection we treat $r$ as a positive integer. Let $Z_1, \ldots, Z_r \stackrel{\mathrm{iid}}{\sim} \mathcal{N}(0,1)$ and $M_r := \max_{j} Z_j$, with
\begin{equation}
\kappa_r := \mathbb{E}[M_r] = \int_{-\infty}^{\infty} z \, r \phi(z) \Phi(z)^{r-1} \, dz,
\label{eq:kappa_r}
\end{equation}
where $\phi$ and $\Phi$ are the standard normal density and CDF. For large $r$, $\kappa_r \sim \sqrt{2 \log r}$.

\begin{theorem}[Barrier-aware Attention load]\label{thm:barrier}
Under the assumptions of Lemma~\ref{lem:stationary} with $\nu > 0$, for every fixed integer $r \ge 1$,
\begin{equation}
\frac{W_{B,r} - B \theta}{\sqrt{B}\, \nu} \;\Rightarrow\; M_r \quad \text{as } B \to \infty,
\label{eq:barrier_clt}
\end{equation}
and
\begin{equation}
\mathbb{E}[W_{B,r}] = B \theta + \sqrt{B}\, \nu\, \kappa_r + o(\sqrt{B}).
\label{eq:barrier_mean}
\end{equation}
If $\nu = 0$, then $W_{B,r} = B \theta$ exactly.
\end{theorem}

The proof, in Appendix~\ref{app:thm_barrier}, combines the multivariate central limit theorem with a uniform-integrability argument bounding $\mathbb{E}[(\max_j X_j^{(B)})^2] \le r$. Equation~\eqref{eq:barrier_mean} gives the barrier-aware replacement for the mean-field Attention load:
$B \theta \;\;\leadsto\;\; B \theta + \sqrt{B}\, \nu\, \kappa_r.$
The relative synchronization overhead is $(\nu/\theta)(\kappa_r/\sqrt{B})$, which grows sublogarithmically with $r$ via $\kappa_r \sim \sqrt{2 \log r}$ and decays as $B^{-1/2}$.

\paragraph{Monte Carlo validation.}
We verify the Gaussian approximation under geometric decode lifetimes (Corollary~\ref{cor:geometric}) with $B = 256$, $\mu_P = 100$, and $\mu_D = 500$ ($\mu_{\mathrm{out}} = 499$). By Corollary~\ref{cor:geometric}, $\theta = \mu_P + \mu_{\mathrm{out}}$ and $\nu^2 = \sigma_P^2 + \mu_{\mathrm{out}}(\mu_{\mathrm{out}} + 1)$, giving predicted overhead $\sqrt{B}\,\nu\,\kappa_r/(B\theta)$. Table~\ref{tab:mc_validation} (Appendix~\ref{app:mc_validation}) compares this against Monte Carlo estimates (50,000 trials per $r$): the CLT prediction matches within 0.5\% for all $r \in \{2, \ldots, 24\}$. The overhead reaches ${\sim}11\%$ at $r = 24$, accounting for the majority of the ${\sim}15\%$ throughput gap observed at large $r$ (Section~\ref{sec:exp}). Importantly, after incorporating this correction into~\eqref{eq:rstar_G}, the simulation-optimal $r^*$ remains at 8, confirming that the mean-field rule~\eqref{eq:rstar_mf} already provides an accurate provisioning recommendation.
\iffalse
We verify the Gaussian approximation under geometric decode lifetimes (Corollary~\ref{cor:geometric}) with $B = 256$, $\mu_P = 100$, $\mu_D = 500$. Each worker's load is $T_j \approx \mathcal{N}(m, s^2)$ with $m = B(\mu_P + \mu_D)$ and $s = \sqrt{B(\sigma_P^2 + \mu_D(\mu_D + 1))}$, giving predicted overhead $\kappa_r s / m$. Table~\ref{tab:mc_validation} (Appendix~\ref{app:mc_validation}) compares this against Monte Carlo estimates ($50{,}000$ trials per $r$): the CLT prediction matches within $0.5\%$ for all $r \in \{2, \ldots, 24\}$. The overhead reaches ${\sim}11\%$ at $r = 24$, accounting for the majority of the ${\sim}15\%$ throughput gap observed at large $r$ (Section~\ref{sec:exp}). Importantly, after incorporating this correction into~\eqref{eq:rstar_G}, the simulation-optimal $r^*$ remains at $8$, confirming that the mean-field rule~\eqref{eq:rstar_mf} already provides an accurate provisioning recommendation.%
\fi
%----------------------------------------------------------
\subsection{Cycle Time Approximations}
\label{sec:cycle_time}
%----------------------------------------------------------

We work with two approximations to the expected cycle time $\bar\tau(B; r)$. Define
\[
\mu_A := \alpha_A B \theta + \beta_A, \qquad
\sigma_A := \alpha_A \sqrt{B}\, \nu, \qquad
G_{B,r} := \max\{\alpha_C rB + \beta_C,\; \alpha_F r B + \beta_F\}.
\]

\noindent \textbf{Mean-field cycle time.} Ignoring the cross-worker barrier corresponds to replacing $W_{B,r}$ by its mean-field value $B\theta$:
\begin{equation}
\tau_{\mathrm{mf}}(B; r) := \max\!\left\{\mu_A,\; \alpha_C rB + \beta_C,\; \alpha_F r B + \beta_F\right\}.
\label{eq:tau_mf}
\end{equation}
This is the natural deterministic surrogate when stochastic imbalance is fully absorbed by an idealized balancing policy ($\nu_{\mathrm{eff}} \to 0$).

\noindent \textbf{Gaussian cycle time.} Retaining the barrier, the leading-order Gaussian approximation to $\bar\tau(B; r)$ is
\begin{equation}
\tau_G(B; r) := G_{B,r} + \sigma_A \int_{z_{B,r}}^{\infty} (m - z_{B,r})\, r \phi(m) \Phi(m)^{r-1}\, dm, \qquad z_{B,r} := \frac{G_{B,r} - \mu_A}{\sigma_A},
\label{eq:tau_G}
\end{equation}
with $\bar\tau(B; r) = \tau_G(B; r) + o(\sqrt{B})$. The integral admits inexpensive numerical quadrature; Appendix~\ref{app:prop_cycle} gives the derivation and a closed-form expression for small $r$.

%----------------------------------------------------------
\subsection{Provisioning Rules}
\label{sec:regimes}
%----------------------------------------------------------

We present a closed-form mean-field rule and a barrier-aware refinement.
\begin{theorem}[Mean-field optimal A/F ratio]\label{thm:meanfield}
Under the mean-field cycle time \eqref{eq:tau_mf}, the throughput is maximized by evaluating the following candidate ratios:
\begin{equation}
\label{eq:rstar_mf}
\boxed{
\begin{gathered}
\displaystyle
r^*_{\mathrm{mf}} \in \Bigg\{\min\!\left\{
\frac{\mu_A-\beta_C}{\alpha_C B},
\frac{\mu_A-\beta_F}{\alpha_F B}
\right\}, \quad
\sqrt{\frac{\beta_C}{\alpha_C B}}, \quad
\sqrt{\frac{\beta_F}{\alpha_F B}},
\quad
\frac{\beta_C-\beta_F}{B(\alpha_F-\alpha_C)} \Bigg\}.
\end{gathered}
}
\end{equation}
Let \(r^*_{\mathrm{mf}}\) be the candidate in \eqref{eq:rstar_mf}
with the largest \(\mathrm{Thr}_{\mathrm{mf}}(B;r)\). Then
\[
\mathrm{Throughput}^*_{\mathrm{mf}}
=
\frac{r^*_{\mathrm{mf}}B}
{(r^*_{\mathrm{mf}}+1)\tau_{\mathrm{mf}}(B;r^*_{\mathrm{mf}})} .
\]
\end{theorem}
The proof is in Appendix~\ref{app:meanfield}. Since communication and FFN
both scale with the aggregate batch \(rB\), communication enters through
\(\tau_{\mathrm{mf}}\) directly, rather than through a separate
\(\theta\)-dependent balance term.
\noindent \textbf{Barrier-aware refinement.} When the cross-worker barrier is retained, the throughput
\begin{equation}
\mathrm{Thr}_G(B; r) := \frac{rB}{(r + 1)\, \tau_G(B; r)}
\label{eq:thr_G}
\end{equation}
is no longer maximized in closed form, because $\tau_G(B; r)$ depends on $r$ both through the FFN term $\alpha_F r B + \beta_F$ in $G_{B,r}$ and through the order-statistic factor $\kappa_r \sim \sqrt{2\log r}$ inside the integral. The barrier-aware optimal ratio is
\begin{equation}
\boxed{\, r^*_G \in \arg\max_{r \in \mathcal{R}} \mathrm{Thr}_G(B; r), \,}
\label{eq:rstar_G}
\end{equation}
where $\mathcal{R}$ is the feasible set of integer fan-ins. This is a one-dimensional analytic optimization evaluable in milliseconds combined with a discrete search over $r$. In the practically common setting where decode lengths approximately follow a geometric distribution—as observed across multiple production LLM traces (Figure~\ref{fig:pd_dist} in Appendix~\ref{append:geo})—the moment statistics admit closed-form expressions, yielding the following specialization.

\begin{corollary}[Geometric decode lifetimes]\label{cor:geometric}
If $P$ and $D$ are independent and $D \sim \mathrm{Geom}(p)$ on $\{1, 2, \ldots\}$, write $\mu_{\mathrm{out}} := (1-p)/p$ for the expected number of generated tokens. Then
\[
\theta = \mu_P + \mu_{\mathrm{out}}, \qquad \nu^2 = \sigma_P^2 + \mu_{\mathrm{out}}(\mu_{\mathrm{out}} + 1),
\]
so the mean-field rule \eqref{eq:rstar_mf} reduces to the form derived in prior analyses of LLM decoding under geometric assumptions, while \eqref{eq:rstar_G} sharpens it with the synchronization correction.
\end{corollary}

\noindent \textbf{Practical recipe.}
Given hardware parameters $(\alpha_A, \beta_A, \alpha_F, \beta_F, \alpha_C, \beta_C)$ and request traces $(P_i, D_i)_{i=1}^{n}$: (i) Estimate $\hat\theta$ and $\hat\nu^2$ from the trace using the nonparametric estimator (Appendix~\ref{app:tracebased}). (ii) Compute $r^*_{\mathrm{mf}}$ in closed form via Theorem \ref{thm:meanfield} as a fast initial recommendation. (iii) Refine via $r^*_G$ from \eqref{eq:rstar_G} when cross-worker imbalance is non-negligible.

\section{Numerical Experiments} \label{sec:exp}

To validate our theoretical framework, we develop an AFD simulator and conduct 
systematic experiments comparing theoretical predictions with simulation results. 
The source code is available at \url{https://anonymous.4open.science/r/AF-release-1C11}.

\subsection{Simulator Design}
The simulator implements a discrete-event, cycle-by-cycle simulation of the $r$A--$1$F topology, 
where each \texttt{Batch} object transitions through a six-state finite state machine 
(Attention $\to$ A2F transfer $\to$ Waiting $\to$ FFN $\to$ F2A transfer $\to$ Waiting $\to$ repeat). 
To maximize resource utilization, the simulator maintains two batches in flight simultaneously: 
while one batch is being processed by the shared FFN server, the Attention instances process 
the other batch in parallel, and vice versa. This interleaved execution allows computation 
on one batch to overlap with communication and processing of the other, effectively hiding 
transfer latencies.

\iffalse
The simulator enforces continuous batching \citep{kwon2023efficient} by maintaining a global FCFS buffer. Whenever a request completes within a batch, the vacated slot is immediately refilled from the buffer before the next decode step begins, ensuring that each Attention instance consistently processes a full microbatch of size $B$.
\fi
%We implement the routing algorithm from {\color{red} Will add}, which sends completed prefill requests to decode workers while guaranteeing minimal load imbalance between attention instances.

\subsection{Experimental Setup}

\noindent \textbf{Latency parameters and configuration.}
We adopt linear latency coefficients calibrated for the DeepSeek-V3 architecture deployed on Huawei Ascend 910C NPUs, obtained via linear regression on real execution traces: $\alpha_A = 0.00165$ cycles/token, $\beta_A = 50$ cycles, $\alpha_F = 0.083$ cycles/request, $\beta_F = 100$ cycles, $\alpha_C = 0.022$ cycles/token, and $\beta_C = 20$ cycles. Appendix~\ref{app:latency} provides the general derivation framework, enabling practitioners to calibrate these parameters for other hardware platforms. Under this configuration, communication latency can be effectively hidden through pipelining ($t_A, t_F > 2 t_C$ across operating regimes), so our analysis focuses on balancing Attention and FFN workloads. We fix batch size $B = 256$, expected decode length $\mu_D = 500$ ($\sigma_D^2=294500$) tokens, and expected prefill length $\mu_P = 100$ ($\sigma_P^2=9900$) tokens. We sweep $r \in \{1, 2, 4, 8, 16, 24, 32\}$ and simulate until $N = 10{,}000$ requests complete per Attention instance. By Theorem \ref{thm:meanfield}, The theoretical optimal A/F ratio is: $r^*_{\mathrm{mf}} \approx 9.3$.

\noindent \textbf{Evaluation Metrics.}
We evaluate the following metrics to assess system performance:

\begin{itemize}

\item \textbf{Stable Throughput per Instance:} To avoid distortion from startup transients and tail effects (e.g., the final requests draining with partially filled batches), we report throughput computed over the first 80\% of completed requests. Specifically, we measure $T_{80\%}$ as the time when $\lceil 0.8N \rceil$ requests have completed, and compute $\text{Throughput}_{\text{per-inst}}^{(80\%)} = \frac{1}{r+1} \cdot \frac{\sum_{i=1}^{\lceil 0.8N \rceil} D_i}{T_{80\%}},$ 
where $D_i$ is the number of output tokens generated for the $i$-th request.

\item \textbf{Time Per Output Token (TPOT)}: The average time to generate a single output token per request, defined as $    \text{TPOT} = \frac{1}{N} \sum_{i=1}^{N} \frac{T_{\text{decode}}^{(i)}}{D_i},$
where $T_{\text{decode}}^{(i)}$ denotes the total decode time for the $i$-th request.
    
\item \textbf{Idle Ratios}: To quantify resource utilization, we measure the fraction of time that each component remains idle waiting for synchronization: $\eta_A = \frac{1}{r} \sum_{i=1}^{r} \frac{t_{\text{idle},A}^{(i)}}{T_{\text{total}}}$, $\eta_F = \frac{t_{\text{idle},F}}{T_{\text{total}}}$,
where $t_{\text{idle},A}^{(i)}$ and $t_{\text{idle},F}$ denote the total idle waiting time of the $i$-th Attention instance and the shared FFN instance, respectively. Lower values indicate better utilization; in an ideally balanced system, both $\eta_A, \eta_F \to 0$.

\end{itemize}
\subsection{Results}

\begin{figure}[t]
\centering
\includegraphics[width=\linewidth]{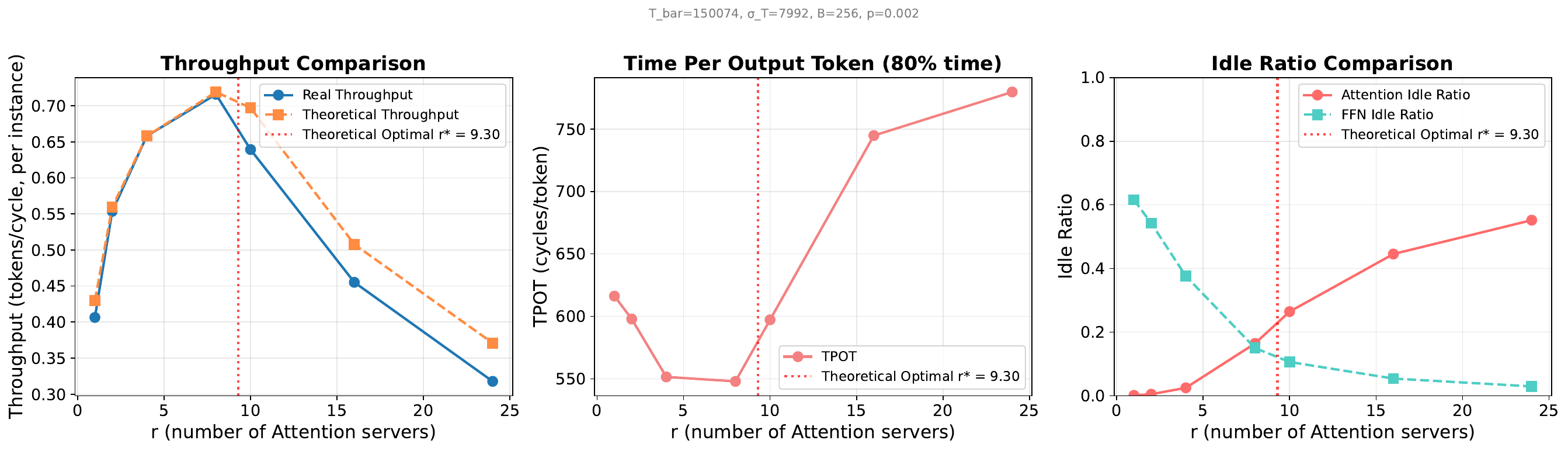}
\caption{Per-instance throughput, TPOT, and idle ratios as functions of A/F ratio $r$ with $B=256$, $\mu_D=500$, $\mu_P=100$. Left: throughput increases with $r$ until reaching the optimal point $r^* \approx 9.3$, after which FFN becomes saturated and throughput per instance decreases. Middle: TPOT decreases and then increases, with the minimum near $r^*$. Right: the crossover point where $\eta_A \approx \eta_F$ indicates the balanced configuration.}
\label{fig:main_results}
\end{figure}

 Figure~\ref{fig:main_results} shows per-instance throughput, TPOT, and idle ratios as functions of $r$. The theoretical optimal $r^*_{\mathrm{mf}} \approx 9.3$ closely matches the simulation-optimal ratio. Moreover, the theoretical throughput curve tracks the simulation results well across all configurations, validating our analytical framework.
A residual gap of up to ${\sim}15\%$ between the mean-field prediction and simulation emerges at large $r$, consistent with the barrier-aware analysis in Section~\ref{sec:barrier}: the synchronization overhead $(\nu/\theta)(\kappa_r/\sqrt{B})$ grows with $r$, reaching ${\sim}11\%$ at $r = 24$ (Table~\ref{tab:mc_validation}), which accounts for the majority of the observed discrepancy. The barrier-aware refinement~\eqref{eq:rstar_G} closes this gap in throughput estimation; in the present configuration, both rules agree on the simulation-optimal $r^*$, indicating that the closed-form mean-field rule already provides accurate provisioning under typical operating conditions.

Figure~\ref{fig:main_results} (right) presents the idle ratios for Attention ($\eta_A$) and FFN ($\eta_F$) instances. When $r$ is small, FFN completes quickly and idles while waiting for Attention outputs ($\eta_F > 60\%$ at $r=1$). As $r$ increases, more Attention instances feed the shared FFN, improving its utilization. %Conversely, when $r$ is large, FFN becomes saturated and Attention instances must wait ($\eta_A > 60\%$ at $r=32$). 

\subsection{Ablation Studies on Workload and Configuration Parameters}

We now investigate how the optimal A/F ratio $r^*$ varies with key system parameters: microbatch size $B$, decode length distribution parameter $p$, and prefill length distribution parameter $q$.

\begin{figure}[t]
    \centering
    \begin{subfigure}{0.48\linewidth}
    \centering
    \includegraphics[width=0.9\linewidth]{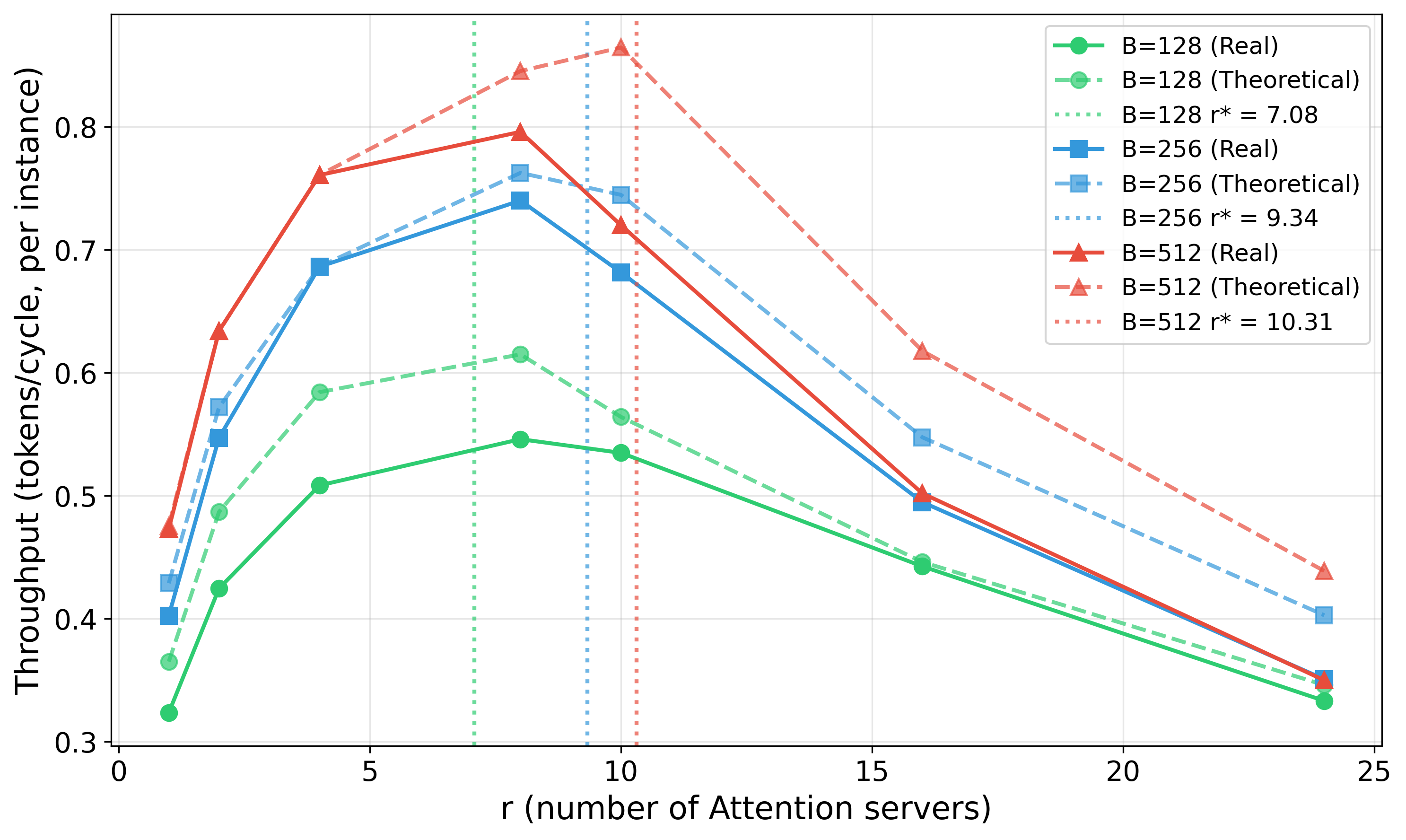}
    \caption{Impact of batch size $B$}
    \label{fig:ablation_B}
    \end{subfigure}\hfill
    \begin{subfigure}{0.48\linewidth}
    \centering
    \includegraphics[width=0.9\linewidth]{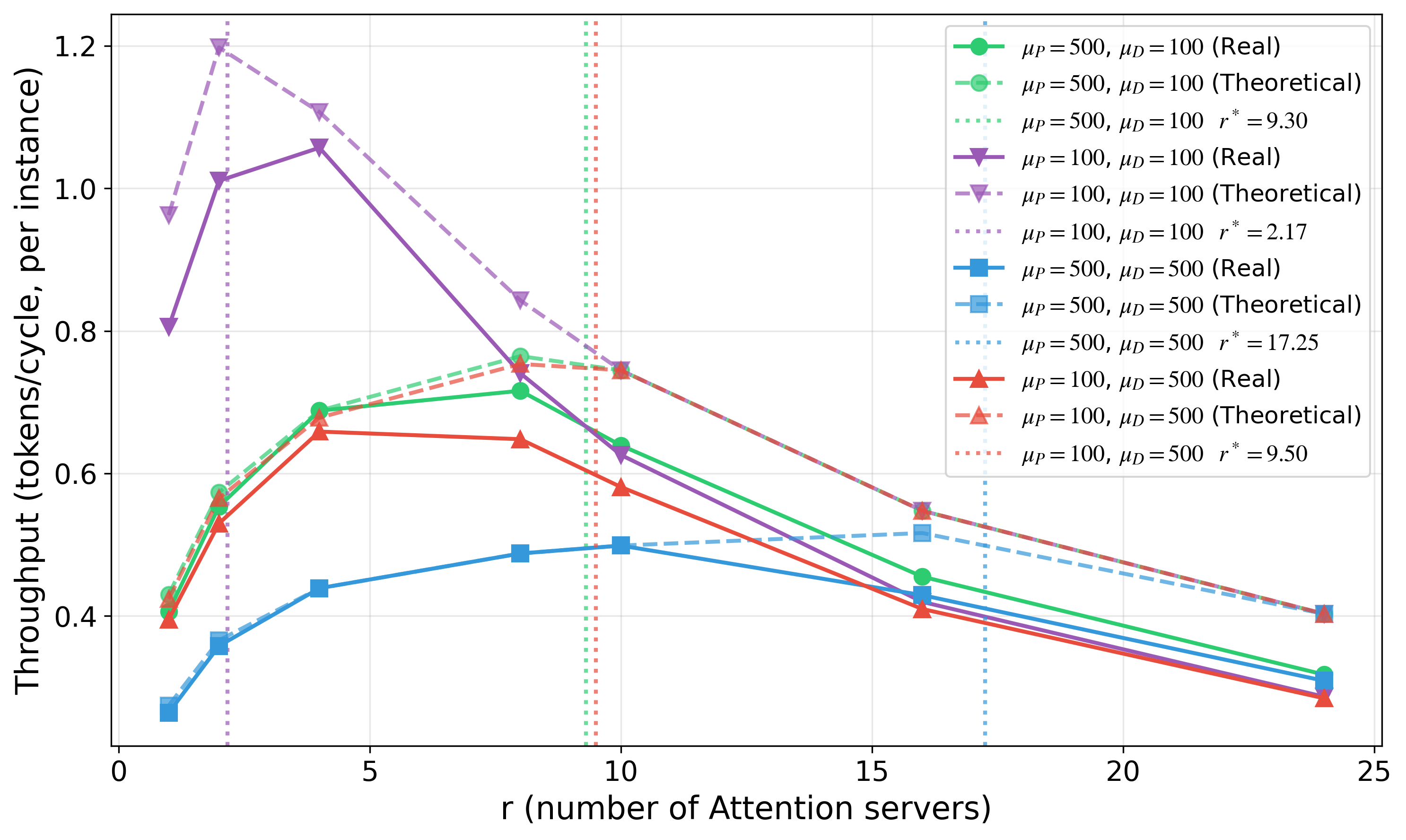}
    \caption{Impact of $\mu_P, \mu_D$}
    \label{fig:ablation_workload}
    \end{subfigure}
    \caption{Ablation studies on per-instance throughput and optimal A/F ratio $r^*$. Solid lines denote simulation results; dashed lines denote theoretical predictions. Vertical dotted lines indicate the theoretical optimal $r^*$ for each configuration.}
    \label{fig:ablation}
\end{figure}

% Ablation Study
\textbf{Ablation on Batch Size $B$.} Figure~\ref{fig:ablation_B} compares throughput across three batch sizes ($B \in \{128, 256, 512\}$) with corresponding theoretical optimal ratios $r^* = \{7.08, 9.34, 10.31\}$. The results demonstrate that larger batch sizes achieve higher peak throughput: batching amortizes fixed overhead costs across more requests, thereby improving hardware utilization. Meanwhile, the optimal $r^*$ increases moderately with $B$, as larger batches require more Attention instances to fully saturate the shared FFN server.

\textbf{Ablation on Workload Distribution.} Figure~\ref{fig:ablation_workload} examines the impact of prefill and decode length distribution. The results reveals that the optimal $r^*$ scales with total context length. Both longer prefills ($\mu_P$) and longer decode sequences ($\mu_D$) increase the total token load in the KV cache, placing greater computational pressure on the Attention side. Consequently, more Attention instances are required to balance the shared FFN server. 

\iffalse
(2) \emph{Longer context lengths reduce peak throughput.} Configurations with larger $\mu_P + \mu_D$ achieve lower maximum throughput at their respective optimal $r^*$. This is because the per-token Attention cost grows with context length---each decode step must attend over a longer KV cache, increasing the marginal computation per output token and limiting overall efficiency.
\fi 
%\subsection{Discussion}

%\begin{ack}
%Add acknowledgments here for the camera-ready version.
%\end{ack}

\bibliographystyle{plainnat}
\bibliography{example_paper}

@inproceedings{zhong2024distserve,
  title={$\{$DistServe$\}$: Disaggregating prefill and decoding for goodput-optimized large language model serving},
  author={Zhong, Yinmin and Liu, Shengyu and Chen, Junda and Hu, Jianbo and Zhu, Yibo and Liu, Xuanzhe and Jin, Xin and Zhang, Hao},
  booktitle={18th USENIX Symposium on Operating Systems Design and Implementation (OSDI 24)},
  pages={193--210},
  year={2024}
}

@inproceedings{patel2024splitwise,
  title={Splitwise: Efficient generative llm inference using phase splitting},
  author={Patel, Pratyush and Choukse, Esha and Zhang, Chaojie and Shah, Aashaka and Goiri, {\'I}{\~n}igo and Maleki, Saeed and Bianchini, Ricardo},
  booktitle={2024 ACM/IEEE 51st Annual International Symposium on Computer Architecture (ISCA)},
  pages={118--132},
  year={2024},
  organization={IEEE}
}

@article{wang2025step,
  title={Step-3 is large yet affordable: Model-system co-design for cost-effective decoding},
  author={Wang, Bin and Wang, Bojun and Wan, Changyi and Huang, Guanzhe and Hu, Hanpeng and Jia, Haonan and Nie, Hao and Li, Mingliang and Chen, Nuo and Chen, Siyu and others},
  journal={arXiv preprint arXiv:2507.19427},
  year={2025}
}

@article{zhu2025megascale,
  title={MegaScale-Infer: Serving Mixture-of-Experts at Scale with Disaggregated Expert Parallelism},
  author={Zhu, Ruidong and Jiang, Ziheng and Jin, Chao and Wu, Peng and Stuardo, Cesar A and Wang, Dongyang and Zhang, Xinlei and Zhou, Huaping and Wei, Haoran and Cheng, Yang and others},
  journal={arXiv preprint arXiv:2504.02263},
  year={2025}
}

@article{zuo2025serving,
  title={Serving Large Language Models on Huawei CloudMatrix384},
  author={Zuo, Pengfei and Lin, Huimin and Deng, Junbo and Zou, Nan and Yang, Xingkun and Diao, Yingyu and Gao, Weifeng and Xu, Ke and Chen, Zhangyu and Lu, Shirui and others},
  journal={arXiv preprint arXiv:2506.12708},
  year={2025}
}

@article{zhang2025janus,
  title={Janus: Disaggregating Attention and Experts for Scalable MoE Inference},
  author={Zhang, Zhexiang and Wang, Ye and Wang, Xiangyu and Zhao, Yumiao and Jiang, Jingzhe and Weng, Qizhen and Shi, Shaohuai and Chen, Yin and Yu, Minchen},
  journal={arXiv preprint arXiv:2512.13525},
  year={2025}
}

@article{brown2020language,
  title={Language models are few-shot learners},
  author={Brown, Tom and Mann, Benjamin and Ryder, Nick and Subbiah, Melanie and Kaplan, Jared D and Dhariwal, Prafulla and Neelakantan, Arvind and Shyam, Pranav and Sastry, Girish and Askell, Amanda and others},
  journal={Advances in neural information processing systems},
  volume={33},
  pages={1877--1901},
  year={2020}
}

@article{chowdhery2023palm,
  title={Palm: Scaling language modeling with pathways},
  author={Chowdhery, Aakanksha and Narang, Sharan and Devlin, Jacob and Bosma, Maarten and Mishra, Gaurav and Roberts, Adam and Barham, Paul and Chung, Hyung Won and Sutton, Charles and Gehrmann, Sebastian and others},
  journal={Journal of Machine Learning Research},
  volume={24},
  number={240},
  pages={1--113},
  year={2023}
}

@article{kaplan2020scaling,
  title={Scaling laws for neural language models},
  author={Kaplan, Jared and McCandlish, Sam and Henighan, Tom and Brown, Tom B and Chess, Benjamin and Child, Rewon and Gray, Scott and Radford, Alec and Wu, Jeffrey and Amodei, Dario},
  journal={arXiv preprint arXiv:2001.08361},
  year={2020}
}

@article{openai2023gpt,
  title={{GPT}-4 technical report. arxiv 2303.08774},
  author={OpenAI},
  volume={2},
  number={5},
  year={2023}
}

@inproceedings{ekelund2025boosting,
  title={Boosting performance of iterative applications on gpus: Kernel batching with cuda graphs},
  author={Ekelund, Jonah and Markidis, Stefano and Peng, Ivy},
  booktitle={2025 33rd Euromicro International Conference on Parallel, Distributed, and Network-Based Processing (PDP)},
  pages={70--77},
  year={2025},
  organization={IEEE}
}

@article{wang2023openchat,
  title={Openchat: Advancing open-source language models with mixed-quality data},
  author={Wang, Guan and Cheng, Sijie and Zhan, Xianyuan and Li, Xiangang and Song, Sen and Liu, Yang},
  journal={arXiv preprint arXiv:2309.11235},
  year={2023}
}

@article{wang2024burstgpt,
  title={Burstgpt: A real-world workload dataset to optimize llm serving systems},
  author={Wang, Yuxin and Chen, Yuhan and Li, Zeyu and Kang, Xueze and Tang, Zhenheng and He, Xin and Guo, Rui and Wang, Xin and Wang, Qiang and Zhou, Amelie Chi and others},
  journal={arXiv preprint arXiv:2401.17644},
  year={2024}
}

@article{zheng2023lmsys,
  title={Lmsys-chat-1m: A large-scale real-world llm conversation dataset},
  author={Zheng, Lianmin and Chiang, Wei-Lin and Sheng, Ying and Li, Tianle and Zhuang, Siyuan and Wu, Zhanghao and Zhuang, Yonghao and Li, Zhuohan and Lin, Zi and Xing, Eric P and others},
  journal={arXiv preprint arXiv:2309.11998},
  year={2023}
}

@article{zhao2024wildchat,
  title={Wildchat: 1m chatgpt interaction logs in the wild},
  author={Zhao, Wenting and Ren, Xiang and Hessel, Jack and Cardie, Claire and Choi, Yejin and Deng, Yuntian},
  journal={arXiv preprint arXiv:2405.01470},
  year={2024}
}

@article{pope2023efficiently,
  title={Efficiently scaling transformer inference},
  author={Pope, Reiner and Douglas, Sholto and Chowdhery, Aakanksha and Devlin, Jacob and Bradbury, James and Heek, Jonathan and Xiao, Kefan and Agrawal, Shivani and Dean, Jeff},
  journal={Proceedings of machine learning and systems},
  volume={5},
  pages={606--624},
  year={2023}
}

@inproceedings{agrawal2024taming,
  title={Taming $\{$Throughput-Latency$\}$ tradeoff in $\{$LLM$\}$ inference with $\{$Sarathi-Serve$\}$},
  author={Agrawal, Amey and Kedia, Nitin and Panwar, Ashish and Mohan, Jayashree and Kwatra, Nipun and Gulavani, Bhargav and Tumanov, Alexey and Ramjee, Ramachandran},
  booktitle={18th USENIX Symposium on Operating Systems Design and Implementation (OSDI 24)},
  pages={117--134},
  year={2024}
}

@inproceedings{kwon2023efficient,
  title={Efficient memory management for large language model serving with pagedattention},
  author={Kwon, Woosuk and Li, Zhuohan and Zhuang, Siyuan and Sheng, Ying and Zheng, Lianmin and Yu, Cody Hao and Gonzalez, Joseph and Zhang, Hao and Stoica, Ion},
  booktitle={Proceedings of the 29th symposium on operating systems principles},
  pages={611--626},
  year={2023}
}

@misc{lienhart2024dissecting,
  author       = {Lienhart, Pierre},
  title        = {{LLM} Inference Series: 5. Dissecting Model Performance},
  howpublished = {Medium},
  year         = {2024},
  month        = mar,
  url          = {https://medium.com/@plienhar/llm-inference-series-5-dissecting-model-performance-6144aa93168f}
}

@article{yuan2024llm,
  title={Llm inference unveiled: Survey and roofline model insights},
  author={Yuan, Zhihang and Shang, Yuzhang and Zhou, Yang and Dong, Zhen and Zhou, Zhe and Xue, Chenhao and Wu, Bingzhe and Li, Zhikai and Gu, Qingyi and Lee, Yong Jae and others},
  journal={arXiv preprint arXiv:2402.16363},
  year={2024}
}

@article{wei2024building,
  title={Building on efficient foundations: Effective training of LLMs with structured feedforward layers},
  author={Wei, Xiuying and Moalla, Skander and Pascanu, Razvan and Gulcehre, Caglar},
  journal={Advances in Neural Information Processing Systems},
  volume={37},
  pages={4689--4717},
  year={2024}
}

@misc{chen2023dissecting,
  author       = {Chen, Lequn},
  title        = {Dissecting Batching Effects in {GPT} Inference},
  howpublished = {Blog post},
  year         = {2023},
  month        = may,
  url          = {https://le.qun.ch/en/blog/2023/05/13/transformer-batching/}
}

@inproceedings{narayanan2021efficient,
  title={Efficient large-scale language model training on gpu clusters using megatron-lm},
  author={Narayanan, Deepak and Shoeybi, Mohammad and Casper, Jared and LeGresley, Patrick and Patwary, Mostofa and Korthikanti, Vijay and Vainbrand, Dmitri and Kashinkunti, Prethvi and Bernauer, Julie and Catanzaro, Bryan and others},
  booktitle={Proceedings of the international conference for high performance computing, networking, storage and analysis},
  pages={1--15},
  year={2021}
}

@article{li2024flash,
  title={Flash Communication: Reducing Tensor Parallelization Bottleneck for Fast Large Language Model Inference},
  author={Li, Qingyuan and Zhang, Bo and Ye, Liang and Zhang, Yifan and Wu, Wei and Sun, Yerui and Ma, Lin and Xie, Yuchen},
  journal={arXiv preprint arXiv:2412.04964},
  year={2024}
}

@article{chen2026universal,
  title={A Universal Load Balancing Principle and Its Application to Large Language Model Serving},
  author={Chen, Zixi and Bu, Tianci and Song, Chendong and Lu, Xin and Ye, Yinyu and Zhou, Zijie},
  journal={arXiv preprint arXiv:2601.17855},
  year={2026}
}

%%%%%%%%%%%%%%%%%%%%%%%%%%%%%%%%%%%%%%%%%%%%%%%%%%%%%%%%%%%%%%%%%%%%%%%%%%%%%%%
% APPENDIX
%%%%%%%%%%%%%%%%%%%%%%%%%%%%%%%%%%%%%%%%%%%%%%%%%%%%%%%%%%%%%%%%%%%%%%%%%%%%%%%
\newpage
\appendix

\section{Supplementary Materials for Section~\ref{sec:optimal}}
\label{append:proofs}

\subsection{Proof of Lemma~\ref{lem:stationary}}
\label{app:lemma_stationary}

We use the discrete-time renewal-reward theorem applied to one decode slot. Each request constitutes one renewal cycle. Cycle $n$ has length $D_n$ steps, during which the slot visits ages $a = 0, 1, \ldots, D_n - 1$. For any measurable function $g(P, a)$ with $\mathbb{E}[\sum_{a=0}^{D-1} |g(P, a)|] < \infty$, the renewal-reward theorem gives the long-run time average
\begin{equation}
\mathbb{E}_\pi[g(P, A)] = \frac{\mathbb{E}\!\left[\sum_{a=0}^{D-1} g(P, a)\right]}{\mathbb{E}[D]},
\label{eq:rrtheorem}
\end{equation}
where $A$ denotes the stationary age and the expectation on the right is taken over a single cycle.

\paragraph{First moment.} Take $g(P, a) = P + a$. Using $\sum_{a=0}^{D-1}(P + a) = DP + D(D-1)/2$,
\[
\theta = \mathbb{E}_\pi[P + A] = \frac{\mathbb{E}[DP + D(D-1)/2]}{\mathbb{E}[D]},
\]
which proves the first identity in \eqref{eq:theta_general}. Condition $\mathbb{E}[D] < \infty$ together with \eqref{eq:moment_assumption} ensures the integrability required by \eqref{eq:rrtheorem}.

\paragraph{Second moment.} Take $g(P, a) = (P + a)^2$. Expanding and using $\sum_{a=0}^{D-1} 1 = D$, $\sum_{a=0}^{D-1} a = D(D-1)/2$, and $\sum_{a=0}^{D-1} a^2 = D(D-1)(2D-1)/6$,
\[
\sum_{a=0}^{D-1}(P + a)^2 = D P^2 + P D(D-1) + \frac{D(D-1)(2D-1)}{6}.
\]
Substituting into \eqref{eq:rrtheorem} proves the second identity in \eqref{eq:theta_general}, and $\nu^2 = \mathbb{E}[Y^2] - \theta^2$.

\paragraph{Independent case.} If $P \perp\!\!\!\perp D$, then $\mathbb{E}[DP]/\mathbb{E}[D] = \mu_P$ and
\[
\frac{\mathbb{E}[D(D-1)]}{2 \mathbb{E}[D]} = \frac{\mu_D^2 + \sigma_D^2 - \mu_D}{2 \mu_D} = \frac{\mu_D - 1}{2} + \frac{\sigma_D^2}{2 \mu_D},
\]
which proves \eqref{eq:theta_indep}. Without independence, $\mathbb{E}[DP]/\mathbb{E}[D] = \mu_P + \mathrm{Cov}(P, D)/\mu_D$, giving the additional covariance term. \hfill $\square$

\subsection{Proof of Theorem~\ref{thm:barrier}}
\label{app:thm_barrier}

Assume $\nu > 0$ (the case $\nu = 0$ is immediate, since $Y_{j,b} = \theta$ a.s.\ implies $W_{B,r} = B\theta$ exactly). Define
\[
X_j^{(B)} := \frac{T_j^{(B)} - B\theta}{\sqrt{B}\, \nu}, \qquad j = 1, \ldots, r.
\]
Each $T_j^{(B)} = \sum_{b=1}^{B} Y_{j,b}$ is a sum of $B$ i.i.d.\ random variables with mean $\theta$ and variance $\nu^2 < \infty$ (by Lemma~\ref{lem:stationary}). By the classical central limit theorem, $X_j^{(B)} \Rightarrow Z_j$ with $Z_j \sim \mathcal{N}(0, 1)$, and the multivariate CLT gives
\[
(X_1^{(B)}, \ldots, X_r^{(B)}) \;\Rightarrow\; (Z_1, \ldots, Z_r),
\]
where $Z_1, \ldots, Z_r$ are independent because the workers are independent. The maximum function $h(x_1, \ldots, x_r) = \max_j x_j$ is continuous, so the continuous mapping theorem yields $\max_j X_j^{(B)} \Rightarrow M_r$. Combined with the identity $W_{B,r} = B\theta + \sqrt{B}\, \nu \max_j X_j^{(B)}$, this proves \eqref{eq:barrier_clt}.

For \eqref{eq:barrier_mean}, observe that $(\max_j X_j^{(B)})^2 \le \sum_{j=1}^{r} (X_j^{(B)})^2$, so
\[
\mathbb{E}\!\left[(\max_j X_j^{(B)})^2\right] \le \sum_{j=1}^{r} \mathbb{E}\!\left[(X_j^{(B)})^2\right] = r,
\]
which establishes uniform integrability of $\{\max_j X_j^{(B)}\}_{B \ge 1}$. Combining uniform integrability with weak convergence yields $\mathbb{E}[\max_j X_j^{(B)}] \to \kappa_r$, and taking expectations in $W_{B,r} = B\theta + \sqrt{B}\, \nu \max_j X_j^{(B)}$ proves \eqref{eq:barrier_mean}. \hfill $\square$

\subsection{Monte Carlo Validation of the Barrier Approximation}
\label{app:mc_validation}

We validate Theorem~\ref{thm:barrier} by comparing the predicted relative synchronization overhead $\kappa_r s / m$ against Monte Carlo (MC) estimates. The experiment uses geometric decode lifetimes (Corollary~\ref{cor:geometric}) with $B = 256$, $\mu_P = 100$, $\mu_D = 500$. Each worker's token load sums $B$ independent slot loads, giving $T_j \approx \mathcal{N}(m, s^2)$ with $m = B(\mu_P + \mu_D)$ and $s = \sqrt{B(\sigma_P^2 + \mu_D(\mu_D + 1))}$. MC estimates use $50{,}000$ independent trials per value of $r$.

\begin{table}[h]
\centering
\caption{Relative synchronization overhead: Monte Carlo vs.\ CLT prediction ($B=256$, $\mu_P=100$, $\mu_D=500$).}\label{tab:mc_validation}
\begin{tabular}{ccc}
\toprule
$r$ & MC overhead & CLT prediction \\
\midrule
2  & 2.98\% & 3.00\% \\
4  & 5.52\% & 5.47\% \\
8  & 7.74\% & 7.57\% \\
12 & 8.88\% & 8.66\% \\
16 & 9.66\% & 9.39\% \\
24 & 11.37\% & 11.01\% \\
\bottomrule
\end{tabular}
\end{table}

The CLT prediction matches the MC estimate within $0.5\%$ across all tested values of $r$, confirming the accuracy of the Gaussian order-statistic approximation.

\subsection{Derivation of the Gaussian Cycle Time \eqref{eq:tau_G}}
\label{app:prop_cycle}

We show that $\bar\tau(B; r) = \tau_G(B; r) + o(\sqrt{B})$, where $\tau_G$ is defined in \eqref{eq:tau_G}.

Let $X_B := \max_j X_j^{(B)}$, so that $\alpha_A W_{B,r} + \beta_A = \mu_A + \sigma_A X_B$. The identity $\max\{a, b\} = b + (a - b)_+$ gives
\[
\max\{\mu_A + \sigma_A X_B,\; G_{B,r}\} = G_{B,r} + \sigma_A (X_B - z_{B,r})_+,
\]
and similarly with $X_B$ replaced by $M_r$. Taking expectations,
\begin{equation}
\bar\tau(B; r) = G_{B,r} + \sigma_A \mathbb{E}[(X_B - z_{B,r})_+].
\label{eq:tau_decomp}
\end{equation}
We claim $\mathbb{E}[(X_B - z_{B,r})_+] - \mathbb{E}[(M_r - z_{B,r})_+] \to 0$ as $B \to \infty$.

\textit{Case 1: $z_{B,r}$ has a finite subsequential limit $z$.} The map $x \mapsto (x - z)_+$ is continuous with linear growth; weak convergence $X_B \Rightarrow M_r$ together with uniform $L^2$-boundedness ($\sup_B \mathbb{E}[X_B^2] \le r$) yields $\mathbb{E}[(X_B - z)_+] \to \mathbb{E}[(M_r - z)_+]$, and Lipschitz continuity in $z$ extends the conclusion to $z_{B,r} \to z$.

\textit{Case 2: $z_{B,r} \to +\infty$.} Cauchy--Schwarz and Markov's inequality give $\mathbb{E}[(X_B - z_{B,r})_+] \le \mathbb{E}[X_B^2] / z_{B,r} \to 0$, and similarly for $M_r$.

\textit{Case 3: $z_{B,r} \to -\infty$.} Write $(X_B - z_{B,r})_+ = (X_B - z_{B,r}) - (X_B - z_{B,r})_-$. The negative part satisfies $(X_B - z_{B,r})_- \le |X_B|\mathbf{1}\{|X_B| > |z_{B,r}|\}$, whose expectation tends to zero by uniform $L^2$-boundedness; the same bound applies to $M_r$, and $\mathbb{E}[X_B] \to \kappa_r$.

In all three cases, the difference vanishes. Substituting into \eqref{eq:tau_decomp} and using the density $f_{M_r}(m) = r \phi(m) \Phi(m)^{r-1}$ to evaluate $\mathbb{E}[(M_r - z_{B,r})_+]$ as the integral in \eqref{eq:tau_G} completes the proof. \hfill $\square$

\paragraph{Closed form for small $r$.} For $r = 1$, $M_1 \sim \mathcal{N}(0, 1)$ and the integral in \eqref{eq:tau_G} reduces to $\phi(z) - z(1 - \Phi(z))$ at $z = z_{B,1}$. For $r = 2$, the integral admits a closed form in terms of $\phi$, $\Phi$, and Owen's $T$-function. For $r \ge 3$, numerical quadrature is straightforward.

\subsection{Proof of Theorem~\ref{thm:meanfield}}
\label{app:meanfield}

Recall
\[
G_{B,r}=\max\{\alpha_CBr+\beta_C,\alpha_FBr+\beta_F\},
\qquad
\tau_{\mathrm{mf}}(B;r)=\max\{\mu_A,G_{B,r}\}.
\]
When \(G_{B,r}\le \mu_A\),
\[
\mathrm{Thr}_{\mathrm{mf}}(B;r)
=
\frac{B}{\mu_A}\frac{r}{r+1},
\]
which is increasing in \(r\). This region ends at
\[
\min\!\left\{
\frac{\mu_A-\beta_C}{\alpha_C B},
\frac{\mu_A-\beta_F}{\alpha_F B}
\right\}.
\]

It remains to consider the region where \(G_{B,r}\) is active. On any
smooth branch \(G_{B,r}=\alpha_iBr+\beta_i\), \(i\in\{C,F\}\),
\[
\frac{d}{dr}\log
\frac{rB}{(r+1)(\alpha_iBr+\beta_i)}
=
\frac{\beta_i-\alpha_iBr^2}
{r(r+1)(\alpha_iBr+\beta_i)}.
\]
Thus the only smooth stationary points are
\[
\sqrt{\frac{\beta_C}{\alpha_C B}},
\qquad
\sqrt{\frac{\beta_F}{\alpha_F B}}.
\]
The only nonsmooth point is the crossing of the communication and FFN
latencies,
\[
\frac{\beta_C-\beta_F}{B(\alpha_F-\alpha_C)}.
\]
Therefore the global optimum must be one of the candidates in
\eqref{eq:rstar_mf}; choosing the one with the largest
\(\mathrm{Thr}_{\mathrm{mf}}(B;r)\) gives \(r^*_{\mathrm{mf}}\). 
Substitution gives the stated optimal throughput. \hfill\(\square\)

\subsection{Trace-Based Nonparametric Estimator}
\label{app:tracebased}

Given a request trace $(P_i, D_i)_{i=1}^{n}$, the framework requires no parametric model. Estimate $\theta$ and the second moment of $Y$ by
\begin{align}
\hat\theta &= \frac{\sum_{i=1}^{n} \big[D_i P_i + D_i(D_i - 1)/2\big]}{\sum_{i=1}^{n} D_i}, \label{eq:theta_hat}\\
\hat q &= \frac{\sum_{i=1}^{n} \big[D_i P_i^2 + P_i D_i(D_i - 1) + D_i(D_i - 1)(2 D_i - 1)/6\big]}{\sum_{i=1}^{n} D_i}, \label{eq:q_hat}
\end{align}
and set $\hat\nu^2 = \hat q - \hat\theta^2$. Both estimators are ratios of i.i.d.\ sums and are strongly consistent under the moment conditions of Lemma~\ref{lem:stationary}. A standard delta-method argument gives $\sqrt{n}$-asymptotic normality under the additional condition that the corresponding cycle-reward variables have finite second moments. Plugging $(\hat\theta, \hat\nu)$ into Theorems~\ref{thm:meanfield} and the barrier-aware rule \eqref{eq:rstar_G} yields the empirical provisioning rule, computable directly from serving logs.

\subsection{Heavy-Tailed Decode Lifetimes}
\label{app:heavy_tail}

The finite-variance Theorem~\ref{thm:barrier} requires $\nu^2 < \infty$. Because the stationary age distribution is length-biased, this is stronger than finite variance of $D$. If $\mathbb{P}(D > x) \sim C x^{-\alpha}$, the stationary age $A$ has tail
\[
\mathbb{P}(A > x) = \frac{\sum_{a > x} \mathbb{P}(D > a)}{\mathbb{E}[D]} \sim \frac{C}{\mathbb{E}[D] (\alpha - 1)}\, x^{-(\alpha - 1)},
\]
shifting the tail exponent from $\alpha$ to $\alpha - 1$. Consequently:
\begin{itemize}
    \item $\alpha > 3$: $\nu^2 < \infty$, and Theorem~\ref{thm:barrier} applies.
    \item $2 < \alpha \le 3$: $\theta < \infty$ but $\nu^2 = \infty$. The Gaussian $\sqrt{B}$ correction is replaced by stable-law fluctuations; if $Y$ lies in the domain of attraction of a $\gamma$-stable law with $\gamma = \alpha - 1$, then $W_{B,r} = B\theta + B^{1/\gamma}\, \mathbb{E}[\max_{1 \le j \le r} S_j] + o(B^{1/\gamma})$ for i.i.d.\ $\gamma$-stable $S_j$.
    \item $\alpha \le 2$: $\theta$ may be infinite; the mean-field token load is undefined.
\end{itemize}
In practice, serving systems impose maximum context and generation lengths, restoring all moments, so condition \eqref{eq:moment_assumption} is satisfied.

\subsection{Evidence of Approximately Geometric Decode Length} \label{append:geo}

\begin{figure}[t]
  \centering
  \includegraphics[width=\columnwidth]{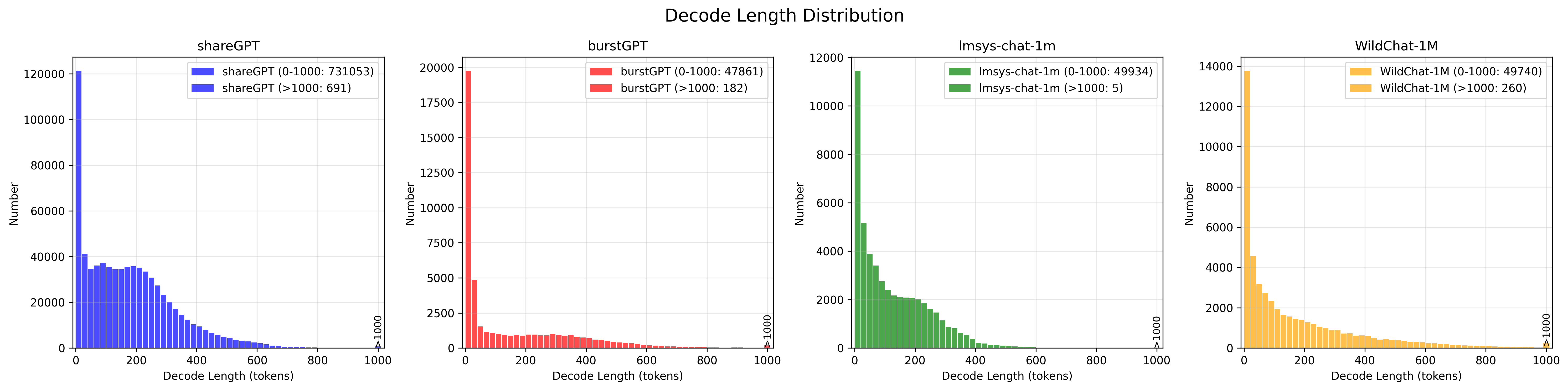}
  \caption{Empirical distributions of decode lengths from production LLM traces \cite{wang2023openchat,wang2024burstgpt,zheng2023lmsys,zhao2024wildchat}. Decode lengths exhibit a geometric (discrete-exponential) pattern.}
  \label{fig:pd_dist}
\end{figure}

Figure \ref{fig:pd_dist} shows empirical distributions of decode lengths from production LLM traces \cite{wang2023openchat,wang2024burstgpt,zheng2023lmsys,zhao2024wildchat}.

%%%%%%%%%%%%%%%%%%%%%%%%%%%%%%%%%%%%%%%%%%%%%%%%%%%%%%%%%%%%%%%%%%%%%%%
% Appendix B: Supplementary Materials for Section 5
% Derivation of Latency Model Parameters based on Ascend 910
%%%%%%%%%%%%%%%%%%%%%%%%%%%%%%%%%%%%%%%%%%%%%%%%%%%%%%%%%%%%%%%%%%%%%%%

\section{Supplementary Materials for Section 5}
\label{app:latency}
This appendix presents the methodology for deriving the latency model parameters $\alpha_A, \beta_A, \alpha_F, \beta_F, \alpha_C, \beta_C$ used in the main text. We demonstrate the derivation process using the DeepSeek-V3 architecture, but the methodology generalizes to other models and hardware platforms.

\paragraph{Note on hardware parameters.} Due to confidentiality constraints, we cannot disclose the explicit Huawei Ascend 910C hardware specifications used in our experiments. Instead, we present the derivation framework using symbolic hardware parameters (summarized in Table~\ref{tab:hw_params}), enabling practitioners to apply this methodology to their own hardware platforms. The final parameter values in Table~\ref{tab:latency_summary} were obtained via linear regression on real execution traces over Huawei Ascend 910C.

\begin{table}[h]
\centering
\caption{Hardware parameters (platform-specific, values not disclosed)}
\label{tab:hw_params}
\begin{tabular}{cl}
\toprule
\textbf{Symbol} & \textbf{Description} \\
\midrule
$\pi_{\mathrm{peak}}$ & Peak INT8 compute throughput (TFLOPS) \\
$\beta_{\mathrm{HBM}}$ & Peak HBM memory bandwidth (TB/s) \\
$\eta_{\mathrm{mem}}$ & Effective memory bandwidth utilization ratio \\
$\eta_{\mathrm{compute}}$ & Effective compute utilization ratio \\
$\beta_{\mathrm{intra}}$ & Intra-node interconnect bandwidth (GB/s) \\
$\beta_{\mathrm{inter}}$ & Inter-node network bandwidth (GB/s) \\
$N_{\mathrm{cards}}$ & Number of accelerator cards in deployment \\
$N_{\mathrm{expert/card}}$ & Number of experts per card \\
\bottomrule
\end{tabular}
\end{table}

%-----------------------------------------------------------------------
\subsection{Model Configuration}
%-----------------------------------------------------------------------

We base our derivations on the DeepSeek-V3 architecture. The model has a hidden size of $H = 7168$ and uses Multi-head Latent Attention (MLA) with a compressed KV cache dimension of $(d_c + d_{\mathrm{rope}}) = 512 + 64 = 576$. For the MoE layers, the expert intermediate dimension is $d_{\mathrm{expert}} = 2048$, with a total of $N_{\mathrm{expert}} = 256$ experts across the system, where each token is routed to $k = 8$ experts. We use Multi-Token Prediction (MTP) with depth 1.

%-----------------------------------------------------------------------
\subsection{MLA Attention (Memory-Bound)}
\label{sec:attn_latency}
%-----------------------------------------------------------------------

During decoding, Attention computation is \emph{memory-bound}, dominated by reading the compressed KV cache from HBM. The compute cost scales with total token load $T = \sum_{b=1}^{B}(s_b + i_b)$.

\subsubsection{Derivation}

\paragraph{Data volume per token.}
With KV compression dimension $(d_c + d_{\mathrm{rope}}) = 576$ and BF16 precision (2 bytes per element):
\begin{equation}
    V_{\mathrm{token}} = (d_c + d_{\mathrm{rope}}) \times 2 = 576 \times 2 = 1152~\text{bytes}.
\end{equation}

\paragraph{Effective HBM bandwidth.}
Accounting for non-coalesced memory access patterns, cache effects, and memory controller overhead, the effective bandwidth is:
\begin{equation}
    \beta_{\mathrm{eff}} = \beta_{\mathrm{HBM}} \times \eta_{\mathrm{mem}}.
\end{equation}

\paragraph{Time per token (slope).}
The per-token memory access time gives the slope parameter:
\begin{equation}
    \alpha_A = \frac{V_{\mathrm{token}}}{\beta_{\mathrm{eff}}} = \frac{(d_c + d_{\mathrm{rope}}) \times 2}{\beta_{\mathrm{HBM}} \times \eta_{\mathrm{mem}}}.
\end{equation}

\paragraph{Fixed overhead (intercept).}
The intercept $\beta_A$ captures the other projections in the attention block, including Query/Key/Value/Output projections in memory-bound status as well as vector operations such as RMSNorm or RoPE. This parameter is best determined empirically via profiling or linear regression on execution traces.

%-----------------------------------------------------------------------
\subsection{MoE FFN (Compute-Bound)}
\label{sec:ffn_latency}
%-----------------------------------------------------------------------

The FFN computation is \emph{compute-bound} when batch size is sufficiently large. Let $B_e$ denote the batch size per expert.

\subsubsection{Derivation}

\paragraph{FLOPs per expert.}
For SwiGLU activation with 3 weight matrices (quantized gate, up, down projections):
\begin{equation}
    \text{FLOPs per expert} = 6 \times H \times d_{\mathrm{expert}} \times B_e = 6 \times 7168 \times 2048 \times B_e \approx 8.81 \times 10^{7} \times B_e.
\end{equation}

\paragraph{Effective compute throughput.}
For MoE workloads with dynamic routing and load imbalance, the effective throughput is:
\begin{equation}
    \pi_{\mathrm{eff}} = \pi_{\mathrm{peak}} \times \eta_{\mathrm{compute}}.
\end{equation}

\paragraph{Compute time per expert.}
\begin{equation}
    t_{\text{expert}}(B_e) = \frac{6 \times H \times d_{\mathrm{expert}} \times B_e}{\pi_{\mathrm{eff}}} = \frac{6 \times H \times d_{\mathrm{expert}}}{\pi_{\mathrm{peak}} \times \eta_{\mathrm{compute}}} \cdot B_e.
\end{equation}

\paragraph{Total FFN time per card.}
With $N_{\mathrm{expert/card}}$ experts per card:
\begin{equation}
    t_{\text{FFN,card}}(B_e) = N_{\mathrm{expert/card}} \times t_{\text{expert}}(B_e) + \beta_{\text{overhead}}.
\end{equation}

\paragraph{Batch size mapping.}
The relationship between global batch $B_F$ and batch per expert $B_e$:
\begin{equation}
    B_e = \frac{B_F \times k \times (1 + \text{MTP depth})}{N_{\mathrm{expert}}} = \frac{B_F \times 8 \times 2}{256} = \frac{B_F}{16}.
\end{equation}

\paragraph{Express in terms of global batch.}
Substituting the batch size mapping, the FFN latency takes the form:
\begin{equation}
    t_F(B_F) = \alpha_F \cdot B_F + \beta_F,
\end{equation}
where:
\begin{equation}
    \alpha_F = \frac{N_{\mathrm{expert/card}} \times 6 \times H \times d_{\mathrm{expert}}}{\pi_{\mathrm{peak}} \times \eta_{\mathrm{compute}}} \times \frac{k \times (1 + \text{MTP depth})}{N_{\mathrm{expert}}}.
\end{equation}

%-----------------------------------------------------------------------
\subsection{Communication}
\label{sec:comm_latency}
%-----------------------------------------------------------------------

Communication involves transferring activations between Attention and FFN instances via high-speed interconnects.

\subsubsection{Derivation}

\paragraph{Data volume per batch.}
For a batch of $B_e$ tokens routed to experts, the activation transfer includes input (factor of 1 with INT8 format) and output (factor of 2 with BF16 format) tokens:
\begin{equation}
    V_{\mathrm{comm}}(B_e) = 3 \times H \times B_e = 3 \times 7168  \times B_e = 21504 \times B_e~\text{bytes}.
\end{equation}

\paragraph{Effective network bandwidth.}
For the AFD topology spanning both intra-node and inter-node communication, the effective aggregated bandwidth depends on the network topology and routing:
\begin{equation}
    \beta_{\mathrm{net}} = f(\beta_{\mathrm{intra}}, \beta_{\mathrm{inter}}, \text{topology}),
\end{equation}
where the function $f$ depends on the specific deployment configuration.

\paragraph{Transfer time.}
\begin{equation}
    t_{\mathrm{comm}}(B_e) = \frac{N_{expert/card} V_{\mathrm{comm}}(B_e)}{\beta_{\mathrm{net}}} + \beta_C = \frac{3 \times H}{\beta_{\mathrm{net}}} \cdot B_e + \beta_C.
\end{equation}

\paragraph{Convert to per-token formulation.}
With batch size mapping $B_e = B_F/16$, the communication latency becomes:
\begin{equation}
    t_C(rB) = \alpha_C \cdot rB + \beta_C,
\end{equation}
where:
\begin{equation}
    \alpha_C = N_{expert/card} \times\frac{3 \times H}{\beta_{\mathrm{net}}} \times \frac{k \times (1 + \text{MTP depth})}{N_{\mathrm{expert}}}.
\end{equation}

%-----------------------------------------------------------------------
\subsection{Summary}
%-----------------------------------------------------------------------

The derivations above provide a general framework for computing latency parameters from hardware specifications and model architecture. The key relationships are:

\begin{itemize}
    \item \textbf{Attention (memory-bound):} $\alpha_A \propto \dfrac{d_c + d_{\mathrm{rope}}}{\beta_{\mathrm{HBM}} \times \eta_{\mathrm{mem}}}$
    
    \item \textbf{FFN (compute-bound):} $\alpha_F \propto \dfrac{H \times d_{\mathrm{expert}}}{\pi_{\mathrm{peak}} \times \eta_{\mathrm{compute}}}$
    
    \item \textbf{Communication:} $\alpha_C \propto \dfrac{H}{\beta_{\mathrm{net}}}$
\end{itemize}

Table~\ref{tab:latency_summary} presents the parameter values used in our experiments. These values were obtained via linear regression on real execution traces collected from Huawei Ascend 910C NPUs, which provides empirical validation of the linear latency models while accounting for system-level effects not captured in first-principles analysis.

\begin{table}[h]
\centering
\caption{Latency parameters obtained via linear regression on execution traces (Huawei Ascend 910C)}
\label{tab:latency_summary}
\begin{tabular}{cccl}
\toprule
\textbf{Equation} & \textbf{Symbol} & \textbf{Value} & \textbf{Unit} \\
\midrule
{Eq.~(2)} 
    & $\alpha_A$ & $1.65 \times 10^{-3}$ & cycles/token  \\
    & $\beta_A$ & 50 & cycles  \\
\midrule
{Eq.~(3)} 
    & $\alpha_F$ & 0.083 & cycles/request  \\
    & $\beta_F$ & 100 & cycles  \\
\midrule
{Eq.~(4)} 
    & $\alpha_C$ & $0.022$ & cycles/token  \\
    & $\beta_C$ & 20 & cycles  \\
\bottomrule
\end{tabular}
\end{table}

\begin{figure}[h]
\centering
\includegraphics[width=0.9\textwidth]{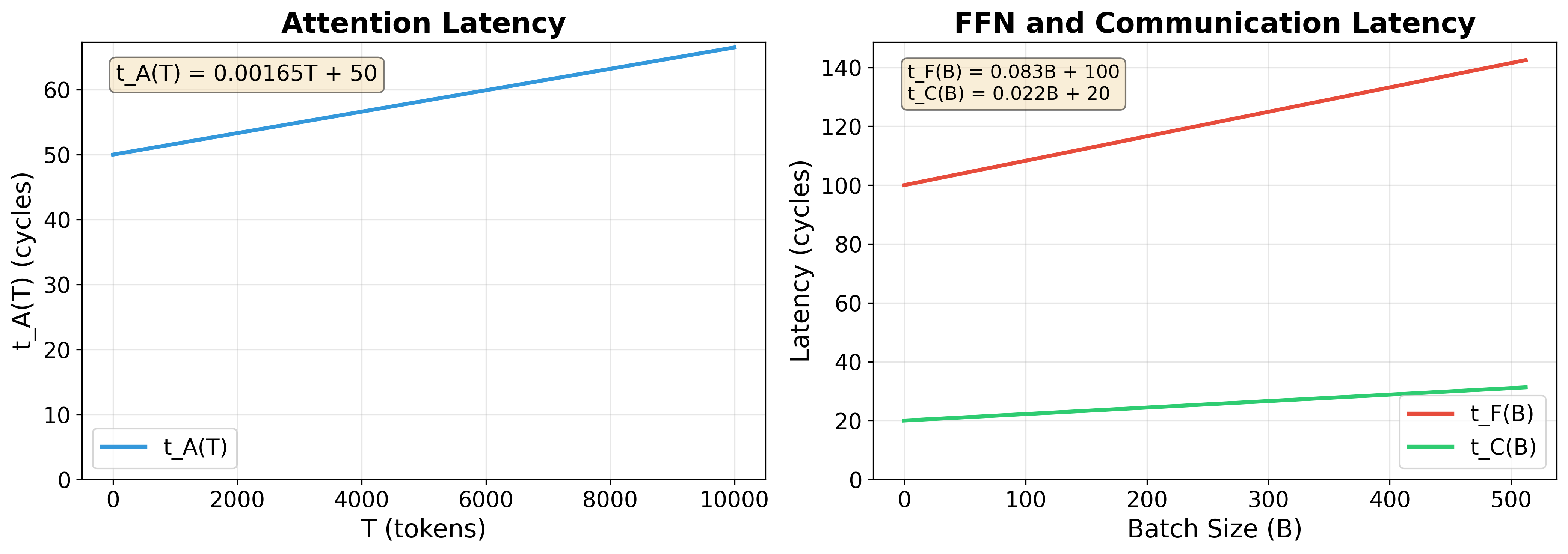}
\caption{Visualization of the latency models. Left: Attention latency $t_A(T)$ scales linearly with total token load $T$. Right: FFN latency $t_F(B)$ and communication latency $t_C(rB)$ as functions of batch size $B$.}
\label{fig:latency_models}
\end{figure}

We visualize the derived latency models in Figure~\ref{fig:latency_models}, which illustrates how Attention, FFN, and communication latencies scale with their respective input sizes under the parameters in Table~\ref{tab:latency_summary}.

\section{Compute resources.}
All simulator experiments reported in Figures~3--4 and Appendix~A.3 are CPU-only and do not require access to NPUs or GPUs once the latency coefficients are fixed. We ran the simulator on [CPU model] with [number] CPU cores and [RAM] memory. A single fan-in sweep for Figure~3 takes approximately 15 minutes, the batch-size and workload ablations in Figure~4 take approximately 20 minutes in total, and reproducing all reported simulator figures takes approximately 3 CPU-hours. The only accelerator-dependent component is the offline latency calibration used to obtain Table~3. These calibration traces were collected on Huawei Ascend 910C NPUs; raw profiling traces and detailed hardware parameters cannot be released due to confidentiality constraints, so reproduction of the paper’s simulator results uses the released fitted coefficients.

\end{document}